\documentclass[11pt]{article}

\usepackage{aaai24}

\usepackage{times}
\usepackage{latexsym}

\usepackage[T1]{fontenc}

\usepackage[utf8]{inputenc}
\usepackage{graphicx}
\usepackage{microtype}

\usepackage{inconsolata}
\usepackage{amsmath}
\usepackage{amssymb}
\usepackage{mathtools}
\usepackage{amsthm}
\usepackage[english]{babel}
\usepackage{booktabs} 
\usepackage{tikz}
\usepackage{tabularx}
\usepackage{natbib} 
\newcommand{\coloredBar}[4]{
  \begin{tikzpicture}
    \definecolor{mycolor}{RGB}{173,216,230}
    \draw[fill=mycolor] (0,0) rectangle (#1*#2,0.2);
    \node at (#3,0.1) {#1#4};
  \end{tikzpicture}
}

\title{Quantifying and Attributing the Hallucination of Large Language Models via Association Analysis}

\author{\textbf{Li Du}, \textbf{Yequan Wang}\footnote{Corresponding Author}, \textbf{Xingrun Xing}, \textbf{Yiqun Yao}, \textbf{Xiang Li}, \textbf{Xin Jiang}, \textbf{Xuezhi Fang}  \\
  Beijing Academy of Artificial Intelligence, Beijing, China \\
        \{ldu, yqyao, lixiang, jiangxin, xzfang\}@baai.ac.cn, \ \ \\
        tshwangyequan@gmail.com, xingxingrun2023@ia.ac.cn  \\
        }

\begin{document}
\maketitle

\def\thefootnote{\arabic{footnote}}
\begin{abstract}
Although demonstrating superb performance on various NLP tasks, large language models (LLMs) still suffer from the hallucination problem, which threatens the reliability of LLMs. To measure the level of hallucination of LLMs, previous works first categorize the hallucination according to the phenomenon similarity, then quantify the proportion that model outputs contain hallucinatory contents.
However, such hallucination rates could easily be distorted by confounders. Moreover, such hallucination rates could not reflect the reasons for the hallucination, as similar hallucinatory phenomena may originate from different sources. To address these issues, we propose to combine the hallucination level quantification and hallucination reason investigation through an association analysis, which builds the relationship between the hallucination rate of LLMs with a set of risk factors. In this way, we are able to observe the hallucination level under each value of each risk factor, examining the contribution and statistical significance of each risk factor, meanwhile excluding the confounding effect of other factors. 
Additionally, by recognizing the risk factors according to a taxonomy of model capability, we reveal a set of potential deficiencies in commonsense memorization, relational reasoning, and instruction following, which may further provide guidance for the pretraining and supervised fine-tuning process of LLMs to mitigate the hallucination. 




\end{abstract}



\section{Introduction}

Large Language Models (LLMs) such as GPT-3 \cite{brown2020language}, PaLM \cite{anil2023palm}, and GPT-4 \cite{bubeck2023sparks} have demonstrated remarkable performance on various natural language tasks \cite{wang2019superglue,rajpurkar2018know,joshi2017triviaqa}. Through the pretraining process upon massive text corpora and the following finetuning process, LLMs are trained to understand natural language, generate highly fluent and realistic responses for the given context, and follow human instructions to complete the demanded tasks \cite{zhao2023survey}. 

Despite the impressive capabilities exhibited by LLMs, it is observed in many cases that the generations of LLMs may be untruthful to contain false \cite{zheng2023does}, illogical \cite{zhong2023can}, fabricated contents \cite{agrawal2023language}, or unfaithful to deviate from human instructions \cite{jones2022capturing,wang2023aligning}. Previous literature concludes this plethora of phenomenons that the LLMs fail to as \emph{hallucination} \cite{ji2023survey}. The hallucination of LLMs arises concerns regarding the trustworthiness of LLMs and limits their application, especially in domains demanding high credibility, such as medicine or financing.


The broad existence of hallucination calls for accurately and comprehensively quantifying the hallucination level of LLMs, to evaluate the reliability of model outputs. However, this task still remains challenging, as the \emph{hallucination} is fundamentally a composition of a set of phenomena\cite{ji2023survey}. Moreover, similar hallucination phenomena may originate from distinct sources \cite{zheng2023does}. Hence, rather than comprehensively quantify the hallucination level of LLMs, primary research makes compromises by focusing on specific subtypes of hallucination (e.g., intrinsic hallucination \cite{ji2023survey}), upon certain tasks, such as text summarization \cite{cao2022hallucinated}, dialogue \cite{shuster2022language,bang2023multitask}, or question answering \cite{anil2023palm,brown2020language}. Then upon a certain task, the hallucination level can be measured using a hallucination rate, i.e., a proportion that the model outputs deviates from the ideal situation. 

However, the hallucination rates may be restricted in faithfully characterizing the hallucination level of LLMs, as it would be easily distorted by confounders, such as the sample distribution of datasets. As for two models with different hallucination rates in two subtypes of instances, by adjusting the proportion of these two types of instances in the dataset, the relative hallucination rate between the two models would change significantly. Hence, a more faithful hallucination quantification should be conducted by managing to control the effects of potential confounders.
Furthermore, a more critical issue is \emph{why} LLMs hallucinate, and how much the risk factors contribute to the hallucination. This is essential for guiding the training process to mitigate the hallucination. However, this issue remains largely unexplored, as most previous research focused on elucidating the origins of hallucinations in domain-specific models. Whereas the differences in training procedure and parameter scales between LLMs and domain-specific models lead to different origins for the hallucination phenomena of the two kinds of model.






To address these issues, we propose to combine the hallucination level quantification of LLMs with the attribution of hallucination through an association analysis. Specifically, we model the relationship between the probability of hallucinating with a set of potential risk factors and recognize the risk factors according to potential impairments in the model's fundamental capabilities. This enables us to compare the hallucination level of LLMs under the same level of each risk factor, to unbiasedly quantify the hallucination level by controlling the effect of potential confounders. Moreover, by observing the sensitivity of the hallucination rate upon these risk factors, and the statistical significance of the risk factors, the contribution of each risk factor can be measured to unveil the reasons for the hallucination of LLMs, so as to facilitate the mitigation of the hallucination.   




Experimental results show that: (1) The LLMs tend to make more on entities or relational knowledge that appears with lower frequency in the corpus, and with more complexity; (2) In the relational reasoning task, the number of arguments and statements significantly positively related to the hallucination rate; (3) The conflicts between human instruction and language modeling significantly related to the hallucination rate.









\section{Related Work}

Before the emergence of large language models, researchers noticed that task-specific generative models may generate nonsensical, unfaithful, or illogical content \cite{filippova2020controlled,maynez2020faithfulness，raunak2021curious}. These research concluded such phenomena as hallucination, then categorized the hallucination according to phenomena similarity and tasks. For example, \cite{ji2023survey} classify all the hallucinations into intrinsic hallucinations and extrinsic hallucinations, where intrinsic hallucinations refer to the generated outputs contradicting the source content; extrinsic hallucinations mean that the generations cannot be verified from the source. \citet{pagnoni2021understanding}, \cite{thomson2020gold} and \citet{li2020slot} explored intrinsic and extrinsic hallucination in abstract summarization, question answering, and dialogue, respectively.   

However, compared to the hallucination of task-specific models, the hallucination of large language models is a broader and more comprehensive concept. 
As multitask model, LLMs are not confined to a single task or specific scenario, which consequently amplifies the potential hallucinations compared to the task-specific models \cite{brown2020language,anil2023palm,bubeck2023sparks}. For example, in the fiction writing task, we may hope the LLM generates unreal content; in counterfactual tasks, we may want the model to generate content that does not conform to facts, which is also ``untruthful''. Additionally, the different training paradigms may bring in different sources of the hallucination phenomena, and further increase the complexity of the hallucination of LLMs \cite{brown2020language,anil2023palm}.  
Considering the complexity of the hallucination of LLMs, pioneer research about the hallucination of LLMs mainly focuses on a certain type of hallucination in confined scenarios, so as to exclude the impact of potential confounding factors. For example, \cite{agrawal2023language,ouyang2022training}  
test whether the model’s ability to separate fact from an adversarially-selected set of incorrect statements using benchmark TruthfulQA \cite{lin2022truthfulqa}. 
\citet{zheng2023does} examine whether the LLMs can memorize, recall and correctly wield commonsense knowledge for answering questions.
\citet{agrawal2023language} test whether the LLMs are aware that they are fabricating non-exist contents by taking the reference generation as an example. \citet{mckenna2023sources} investigate the source of hallucination in the natural language inference task. However, the hallucination rate may easily be influenced by confounders such as data distribution, while in the hallucination attributing works, the specific contributions for the sources of hallucinations are not yet quantified. Moreover, the multitasking nature of LLMs would make it rather laborious to investigate the hallucination of LLMs in each task. To address these issues, in this paper, we propose to analyze the association between the hallucination level of LLMs with a set of risk factors about deficiencies in the model capability, so that the hallucination level quantification and hallucination attributing can be conducted simultaneously.


\section{Methodology}

\subsection{Formalization of Hallucination}
Before describing our methodology, we first formalize the hallucination of LLM to elucidate the scope of this research. Previous literature defines the model hallucination as generating content unfaithful, untruthful, illogical, or contradictory to the context \cite{filippova2020controlled,maynez2020faithfulness，raunak2021curious,ji2023survey}. However, in certain situations such as in novel writing or counterfactual generation tasks, the LLMs might be required to generate content that disobeys the previous hallucination definition. This makes the traditional definition of model hallucination inappropriate for LLMs.

The ultimate goal of the LLM is to follow human instructions, to complete the demanding of human beings, with respect to the historical context. From this perspective, in this paper, we define the hallucination as \emph{generations of the model that violate the human instructions}:
\begin{equation}
    P(Y^*|C, I, T) \neq  P(Y|C, I, T)
\end{equation}
where $Y$ is the ideal output, $Y^*$ is the actual model generation, $C$ is the context, e.g., dialogue history; $T$ is the demanding task, e.g., text summarization, and $I$ denotes the human instruction.


\begin{figure}
    \centering
    \includegraphics[width=0.49\textwidth]{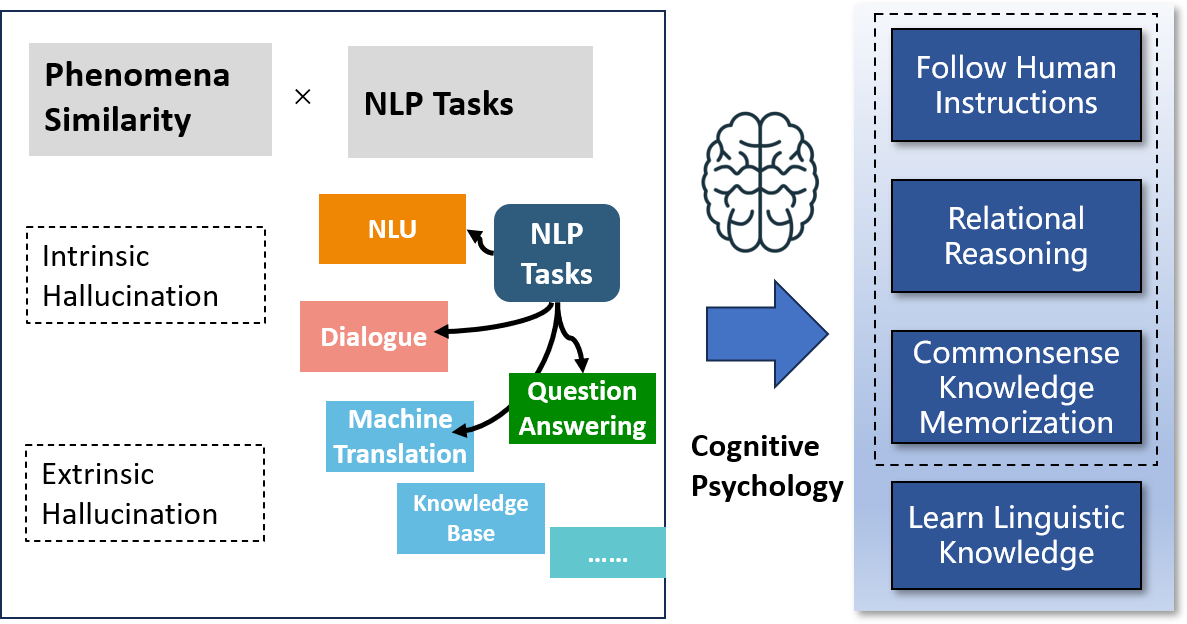}
    \caption{Relationship between the hallucination categorization taxonomy in previous research with our categorization.}
    \label{fig:taxo}
\end{figure}

\subsection{Association Analysis-Based Hallucination Level Quantification}

Previous works quantify the hallucination level of an LLM $\mathcal{M}$ using a single index $P(h_{\mathcal{M}})$, denoting the probability that the model generations contain hallucinatory content upon a certain type of task. In this paper, we propose to quantify the hallucination level through a sensitivity, which investigates how the hallucination level of $\mathcal{M}$ varies with the value of a set of risk factors, i.e., $P(h_{\mathcal{M}})=f_{\mathcal{M}}(R_1, \dots, R_n, C_1, \dots, C_m)$, where $f_{\mathcal{M}}$ is a function, $R_i$ is a risk factor, $C_j$ is the $j$th confounder. So that using $f_{\mathcal{M}}$, we can: (1) Obtain model hallucination level $P(h_{\mathcal{M}})$ with each value of $R_i$, meanwhile control the effects other risk factors and confounders to increase the unbiasedness; (2) Explain the source of the hallucination by examining the sensitivity of $P(h_{\mathcal{M}})$ over $R_i$, then offer guidance for the further mitigation of hallucinations. Additionally, given a set of LLMs $\{\mathcal{M}_1, \dots, \mathcal{M}_M \}$, corresponding sensitivity analysis models $\{f_1, \dots, f_M \}$ can be fitted for each LLM. Then we can compare the hallucination level of different models with the value of risk factors aligned, and the effects of confounders controlled. 

Without generality, we set $f_{\mathcal{M}}$ as a linear logistic regression function:
\begin{equation}
    Y_{k, \mathcal{M}}=\frac{1}{1+\mathrm{exp}^{\beta_{0, \mathcal{M}}+\sum_i \beta_{i, \mathcal{M}} r_{i,k} + \sum_j \gamma_{i, \mathcal{M}} c_{i,k}}}
\end{equation}
where $Y_{k, \mathcal{M}}$ is a 0/1 label denoting whether the generation of $\mathcal{M}$ is hallucinatory given an input $X_k$; $\beta_{0, \mathcal{M}}$ is an interception term, $r_{i,k}$ and $c_{j,k} $is the $k$th value of the $i$th risk factor and the $j$th confounder, respectively.

With such formalization, the regression coefficient $\beta_{i, \mathcal{M}}$ can be explained as how much fold the risk of hallucination changes if the level of risk factor changes 1 unit, i.e, $p(Y_i|r_{i,k}+1)/p(Y_i|r_{i,k})=\mathrm{exp}(\beta_{i, \mathcal{M}})$.


\begin{figure*}
    \centering
    \includegraphics[width=0.99\textwidth]{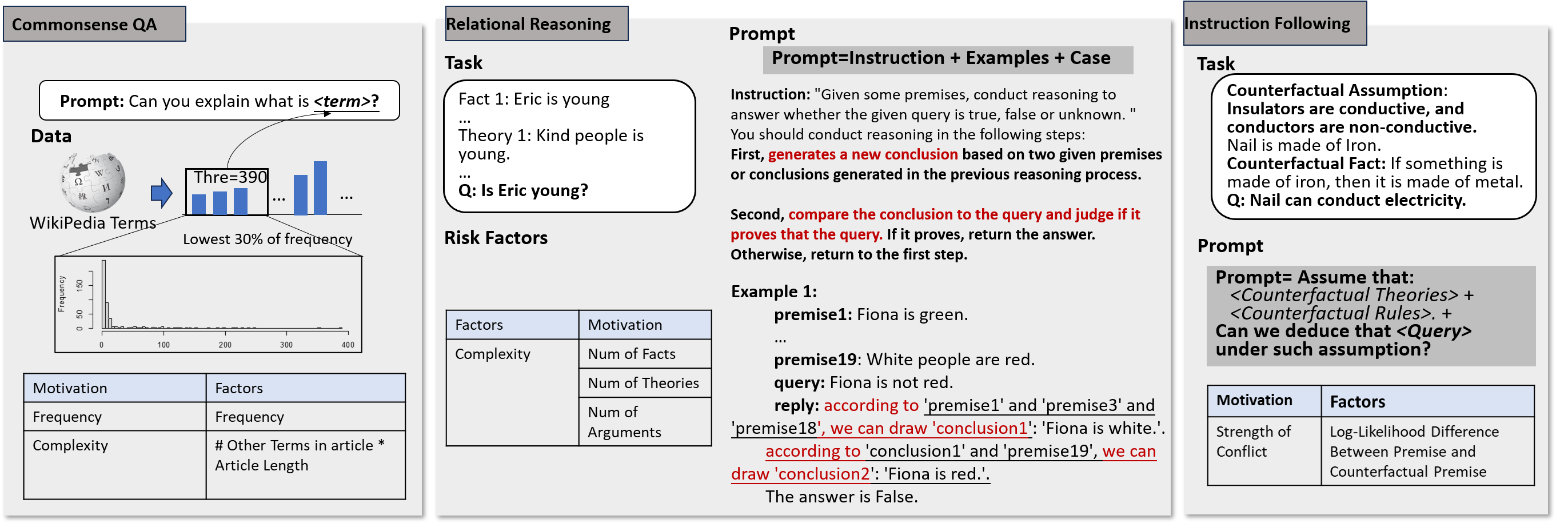}
    \caption{Illustration for the form and prompt of three tasks for quantifying the hallucination level and risk factors for attributing the source of hallucinations.}
    \label{fig:tasks}
\end{figure*}

\subsection{Model Capability Based Risk Factors Identification}

A critical issue of our approach is how to identify the potential risk factors more comprehensively.
Ideally, the risk factors should be related to reasons why the model fails to generate outputs satisfying human demands. Hence, we choose to find the risk factors from the perspective of potential deficiency in the model's fundamental capabilities. To this end, a taxonomy of the model's pivotal capability is necessary, so that under such taxonomy, we dive into the training process of LLMs to recognize the potential risk factors that lead to the deficiency of each subtype of model capabilities.

\subsubsection{Model Capability Deficiency based Hallucination Taxonomy}

Most of the previous research categorizes the hallucination into subtypes according to phenomena similarity, then assesses the level of subtypes of hallucination upon specific tasks, such as question answering, or language understanding. However, such a taxonomy fails to reveal the reasons for hallucinations.  
Additionally, one prominent characteristic of LLMs is that they are multitasking models that can accommodate various NLP tasks \cite{brown2020language,anil2023palm,bubeck2023sparks}. It would be rather laborious to enumerate all possible datasets and scenarios.

Hence, instead of categorizing the hallucination using the taxonomy of tasks, we resort to the deficiency in model capability, in other words, why the LLM fails to complete the demanded task. 
However, there is currently no universally accepted theory to categorize the various capabilities of large LLMs. Considering that the LLMs can achieve impressive performance on certain cognitive reasoning tasks, we categorize their capabilities into the following levels based on cognitive psychology theories \cite{sternberg2012cognitive,barsalou2014cognitive}, and the deficiency in each level of model capability would lead to hallucination. Figure~\ref{fig:taxo} shows our categorization for the hallucination of LLMs, and the relationship between the previous works. In specific: 

(1) Learn linguistic knowledge, including lexical, syntax, morphology, and pragmatics knowledge. Which is the foundation of understanding natural language;

(2) Memorize and recall commonsense knowledge (such as the definition of entities, the relationship between entities, and the relationship between events) for understanding the language. Commonsense knowledge supports the commonsense reasoning process, to allow the model to understand the material world and human society. Cognitive psychology theory concludes such ability as Crystallized Intelligence \cite{ziegler2012openness,cattell1963theory};

(3) Understand abstract relationships and make inferences. Which is the foundation of high-level cognitive processes, such as planning, problem-solving, and decision-making \cite{miller2009executive,alexander2016relational}. Such ability is called Fluid Intelligence in cognitive psychology \cite{ziegler2012openness,cattell1963theory}; 

(4) Understand and follow human instructions.

Through the pretraining process on the large-scale corpus, the LLMs have demonstrated impressive performance upon various linguistic tasks, such as syntactic parsing. Hence, in this paper, we focus on testing the relationship between the model hallucination with the (2) (3), and (4) levels of ability. Based on linguistic knowledge, commonsense knowledge, and relationship modeling ability, LLMs are able to model language. On this basis, models learn to complete tasks under human instruction. The composition of these fundamental abilities can cover a wide span of NLP tasks. Hence, we design tasks and recognize risk factors to probe the deficiency of model capabilities.

\subsubsection{Task Setting}

However, it would be rather challenging to construct a task that can simultaneously probe all the capabilities abovementioned. Hence, in this paper, we separately quantify the hallucination brought by different subtypes of model capability deficiency using a commonsense QA task, a relational reasoning task, and a counterfactual commonsense reasoning task. Figure~\ref{fig:tasks} shows the tasks and corresponding prompts.


\textbf{Commonsense QA Task}

Among various kinds of commonsense knowledge, we focus on investigating the hallucination level about the definition of entities and abstract concepts. These two kinds of knowledge stand as representatives of two major kinds of commonsense knowledge, factual knowledge. As other kinds of factual knowledge and relational knowledge, such as event and event relationship knowledge share a similar learning process, the conclusion on entities/concepts and their relationships would also be applicable to other kinds of factual knowledge and relational knowledge.


To manage to avoid the influence of dataset biases in the existing selective commonsense QA benchmarks \cite{clark2019don,wu2020improving} and facilitate further risk factor identification, we choose to construct a dataset de novo. Based on the Wikitext dataset, we obtain the concepts and entities terms by collecting the title of the articles in the Wikitext dataset. Among all terms, we mainly focus on the long-tail terms. Hence, we calculate the frequency of each term and construct samples based on the terms with the lowest 10\% frequency. The frequency distribution is shown in Figure~\ref{fig:tasks}. Then the sampled terms are filled into the templates to form prompts and submitted to LLMs for obtaining the description or explanation of the terms. Hence, the influence of dataset biases can be managed to avoid by the generative task setting and unified prompt. Moreover, the questions are about single terms rather than complex questions such as that in the TruthfulQA dataset, which facilitates the calculating of statistical characteristics. 
After that, the answers of LLMs are assigned to human annotators to examine their correctness. 

\textbf{Relational Reasoning}

Previously a set of natural language-based relational reasoning tasks have been proposed. However, model performance upon these tasks would also be influenced by confounders, such as whether the model possesses the necessary background knowledge for understanding the statement and the relationship. Hence, To get rid of the effect of confounders, we conduct experiments on the natural language satisfiability task (NLSat) \cite{richardson2022pushing}. Figure~\ref{fig:tasks} shows an example of this task, which requires the model to perform deductive reasoning to derive a conclusion based on a set of systematically constructed short-sentence-based, explicitly stated rules and facts. The rules and facts can be regarded as logical statements expressed by natural language, without involving any additional world knowledge. Hence the potential confounding effect of language and background knowledge can be excluded. We construct the instances by controlling the number of arguments, predicates, statements, and theories, then observe the hallucination rate of LLMs upon these samples. Moreover, note that the results can be deduced by pure symbolic calculation. This enables us to examine the correctness of the LLM's reasoning process by comparing it with the symbolic calculation process.

\textbf{Instruction Following}

Multiple aspects of reasons can make the model fail to follow human instructions, such as (1) the task being too complex for a model to accomplish, (2) the instructions are not clearly enough to describe the task, or (3) the model is inherently cannot accomplish certain tasks. However, (1) and (2) are not brought about by the inherent flaws of the model. Hence, in this paper, we mainly focus on why an LLM inherently cannot accomplish a certain task. 

Previous analyses show that the transformer can approximate any seq2seq function \cite{yun2019transformers,edelman2022inductive}. Hence, the inductive bias of LLM would not be the restriction. We argue that an LLM inherently cannot accomplish a certain task that may originate from the conflict between the pretraining stage and the supervised finetuning (SFT) stage.  
Note that, in the pretraining stage, the task can be defined as $T=$Generating the subsequent sequence of $S$. With regard to $T$,  the object in the pretraining stage may conflict with that in the tuning stage, i.e., $T_{\text{pretraining}} \neq  T_{\text{finetuning}}$. For example, in the pretraining stage, we hope the LLM to learn to generate the given text, while in the finetuning stage, the model may be required to generate counterfactual texts, such as fiction. This conflict may distort the outputs to make them deviate from human instructions. 
This implies that the training objective of LLM in the pre-training stage may not necessarily align with that in the human alignment stage. This conflict may lead to potential hallucinations. Interestingly, such phenomena are also observed in the cognitive process of human beings, which was concluded as the conflict between the subconscious and conscious, or conflict between the perception system and cognition system. 

To detect the contribution of such conflict to the model hallucination level, we propose a \textbf{c}ounterfactual \textbf{N}atural \textbf{L}anguage \textbf{I}nference (c-NLI) task. Specifically, the same as the traditional NLI task, this task requires the model to predict a label with a value of ``entailment'' or ``contradictory'' for a set of given statements. While as Figure~\ref{fig:tasks} shows, the statements in the c-NLI task are all counterfactual, such as ``Insulator can conduct electricity''. In such a situation, the object in the SFT stage would drive the model to complete the reasoning process with the counterfactual setting, while in the pretrain stage, the LLMs are trained to model language by assigning counterfactual generations a small likelihood. Hence, the abovementioned conflict would occur in such instances.  

By constructing the dataset based on commonsense knowledge largely covered by the corpus for training LLMs and controlling the length of each theory and statement, the confounders can be excluded.
We provide the details of constructing the c-NLI dataset in the Appendix.

\subsubsection{Risk Factor Recognition }

Based on the categorization of the fundamental ability of LLMs, we dive into the training process of LLMs to find the risk factors that lead to the deficiency of these abilities and further lead to hallucinations. 


\textbf{Commonsense Memorization}

With such an object, the LLM is trained to ``memorize'' the training data \cite{bang2023multitask,tirumala2022memorization}. Hence, intuitively, whether the model could memorize a piece of commonsense knowledge would be decided by two aspects: 
(1) The frequency of the commonsense knowledge appearing in the corpus; (2) The complexity of the commonsense knowledge: the more complex the commonsense knowledge is, the hard would it be for the model to completely remember the commonsense knowledge and precisely recall the commonsense knowledge.

Given the vast scale of the corpus used for training the model and the diversity of the expressions, it would be challenging to accurately count how many times a particular entity or concept is defined within the entire corpus. Hence, rather than focusing on all possible entities 
 or concepts in the whole corpus, we focus on the terms of Wikipedia. Since the entities or concepts and their relationships are mainly described in the Wikipedia Corpus, we approximate the frequency of the terms of Wikipedia using count their frequency in the Wikipedia corpus. 


Inspired by cognitive psychology, we measure the complexity of an entity/concept using descriptive complexity, i.e., the complexity of an entity or concept can be measured using the effort that is required to describe something \cite{immerman2012descriptive,wimsatt1972complexity,grohe2017descriptive}. In specific, we employ: (1) The length of the corresponding article of this term in Wikipedia, (2) How many other terms appear in the article, (3) The sum of the length of the corresponding article for other terms that appear in the article of one term. 


\textbf{Relationship Reasoning}

Formally, the relationship reasoning process can be abstracted as understanding the relationships between a set of arguments and then making an inference. Where the relationships are described by a set of theories and statements, and each statement and theory is composed of a predicate and several arguments. Hence, the main challenge for correctly conducting the reasoning process to draw a conclusion lies in the complexity of the relational reasoning process, which can be measured using: (1) The number of facts; (2) The number of theories; (3) The number of arguments.


\textbf{Following Human Instructions}

Given the instances of c-NLI, we measure the strength of potential conflict between the objective of two training stages of LLMs using the log-likelihood decrease by changing the statements into the corresponding counterfactual forms, for example, as shown in Figure~\ref{fig:tasks}, changing the assumption from ``\emph{conductors} are \emph{conductive}, \emph{insulators} are \emph{non-conductive}'' to ``\emph{insulators} are \emph{conductive}, \emph{conductors} are \emph{non-conductive}''. As the more decrease of the log-likelihood, the more likely the model would fail to conduct inference by assuming the correctness of the counterfactual statements. 


Then given the datasets, we obtain corresponding generations of LLMs and make judgments about whether a generation contains hallucinatory contents through human annotation.





\section{Experimental Setup}

\subsection{Models}

We examine the hallucination rate of the following OpenAI models:

(1) \textbf{GPT-3} \cite{brown2020language} is a generative language model pretrained upon text corpus with 175B parameters. 

(2) \textbf{text-davinci-001} is a GPT-3.5 series model, which is pretrained upon both text and code corpus, then aligned with human beings through supervised finetuning and reinforcement learning from human feedback (RLHF) to generate longer outputs with consistent instruction-following.

(2) \textbf{text-davinci-003} is further optimized based on InstructGPT to deliver clearer, more engaging, and more compelling content with enhanced commonsense knowledge and relational modeling ability.

(3) \textbf{GPT-3.5-Turbo} is the most capable GPT-3.5 model optimized for chat.

(4) \textbf{GPT-4}  is a large multimodal model with broader general knowledge and advanced reasoning capabilities. Like GPT-3.5-turbo, GPT-4 is optimized for chat but works well for traditional NLP tasks.

\subsection{Evaluation Dataset Construction}

We build the commonsense QA dataset based on the WikiText corpus. We calculate the frequency of each term, a total of 200 terms are sampled from the lowest 30\% frequency.

In the NLSat task, following \cite{richardson2022pushing}, we conduct experiments on the test set of the Ruletaker-depth-3ext dataset (abbreviated as Ruletaker). The experiments are conducted with the few-shot setting, for each instance, we randomly choose n=3,4,5 examples from the corresponding training set to present to the model in the prompt.

We build the counterfactual natural language inference dataset based on the eQASC dataset \cite{jhamtani2020learning}, which provides a set of human-annotated reasoning chains about commonsense knowledge, and such commonsense knowledge is largely covered by the corpus of LLMs. In practice, we obtain the counterfactual statements using the GPT4 API, then check the correctness of automated generated counterfactual statements by human annotation. More details about the data construction process are provided in the Appendix. Totally 150 instances are generated to test the model hallucination rate. 

As the hidden states of Open-AI LLMs are unknown, we obtain the log-likelihood of the statements and counterfactual assumptions using RoBERTa-Large \cite{liu2019roberta}.

\subsection{Human Annotation}

In the commonsense QA and the c-NLI experiment, the outputs of LLMs are submitted to human annotators to check their correctness. We assign each generated output to two annotators simultaneously. Only both of the annotators deem that there is no factual error (for commonsense QA) or reasoning mistake (for c-NLI), then the output is taken as non-hallucinated. 

\subsection{Hyperparameters}
In all experiments, the hyperparameters are set as default values, with Top-P=1 and temperature = 1.

\section{Results and Discussion}

\subsection{Commonsense Knowledge Memorization}

\begin{table}[]
    \centering
    \small
    \begin{tabular}{c|c|c}
    \toprule
       Model & $\beta_{\text{freq}}$ & $\beta_{\text{complexity}}$   \\
    \midrule
       GPT-3  & \multicolumn{1}{c}{\parbox{2cm}{\coloredBar{-0.22}{4}{-0.5}{}}} & \multicolumn{1}{c}{\parbox{2cm}{\coloredBar{0.23}{6}{1}{}}}  \\
       Text-davinci-001 & \multicolumn{1}{c}{\parbox{2cm}{\coloredBar{-0.36}{4}{-1}{$^{.}$}}} & \multicolumn{1}{c}{\parbox{2cm}{\coloredBar{0.35}{6}{1}{}}}  \\
       Text-davinci-003  & \multicolumn{1}{c}{\parbox{2cm}{\coloredBar{-0.35}{4}{-1}{$^{.}$}}} & \multicolumn{1}{c}{\parbox{2cm}{\coloredBar{0.19}{6}{1}{}}}  \\
       GPT-3.5-turbo  & \multicolumn{1}{c}{\parbox{2cm}{\coloredBar{-0.53}{4}{-1}{$^{**}$}}} & \multicolumn{1}{c}{\parbox{2cm}{\coloredBar{0.031}{6}{1}{}}} \\
       GPT-4  &\multicolumn{1}{c}{\parbox{2cm}{\coloredBar{-0.53}{4}{-1}{$^{*}$}}}  & \multicolumn{1}{c}{\parbox{2cm}{\coloredBar{0.26}{6}{1}{}}} \\
    \bottomrule
    \end{tabular}
    \caption{Regression coefficients for different OpenAI LLMs on commonsense memorization task. Statistical significance indicators: ·:p<0.1; *:p<0.05; **:p<0.01; ***: p<0.001.}
    \label{tab:beta-cqa}
\end{table}

\begin{figure}
    \centering
    \includegraphics[width=0.33\textwidth]{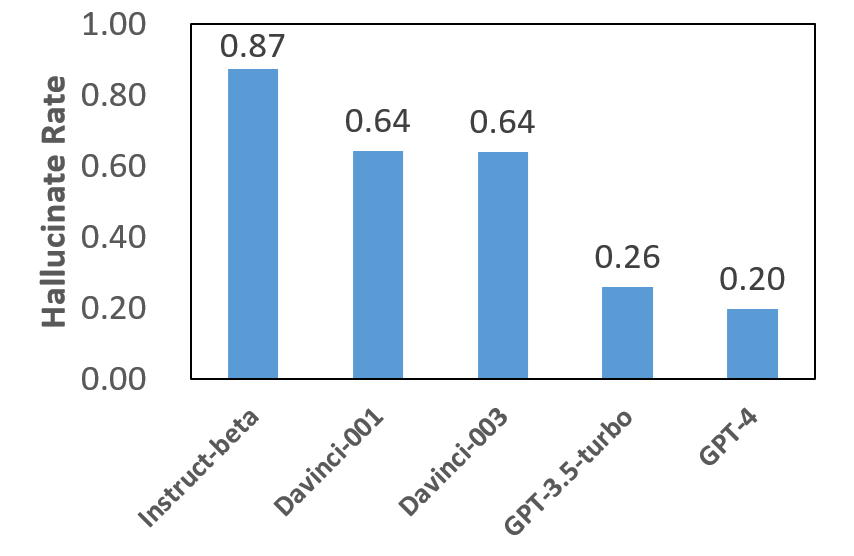}
    \caption{Hallucination rate of OpenAI LLMs on the commonsense memorization task.}
    \label{fig:cqa}
\end{figure}

Table~\ref{tab:beta-cqa} and Figure~\ref{fig:cqa} summarize the relationship between the hallucination rate and the value of risk factors in commonsense knowledge memorization. From which we observe that, 

(1) State-of-the-Art LLMs are still suffering from memorizing long-tailed commonsense knowledge in the training corpus. For example, GPT-4 has a hallucination rate of 19.5\% on explaining Wikipedia terms with 30\% lowest frequency. 

(2) The frequency and the length of the terms are statistically significantly related to the hallucination rate. This pattern shares with the four kinds of LLMs. This suggests that controlling other factors, the higher the frequency of a certain entity or concept in the corpus, the easier it is to be remembered by the LLMs. However, as the complexity of the entity or concept, the model is more likely to generate hallucinatory outputs. This indicates that, during the training process, adding the weight of the instances with lower frequency and higher complexity would decrease the hallucination rate. 

(3) Comparing the average hallucination rate with the regression coefficients suggest that, in general, a model with a lower hallucination level would also be more insensitive to the change in the risk factors. 


\subsection{Relational Reasoning}

 \begin{figure}
    \centering
    \includegraphics[width=0.33\textwidth]{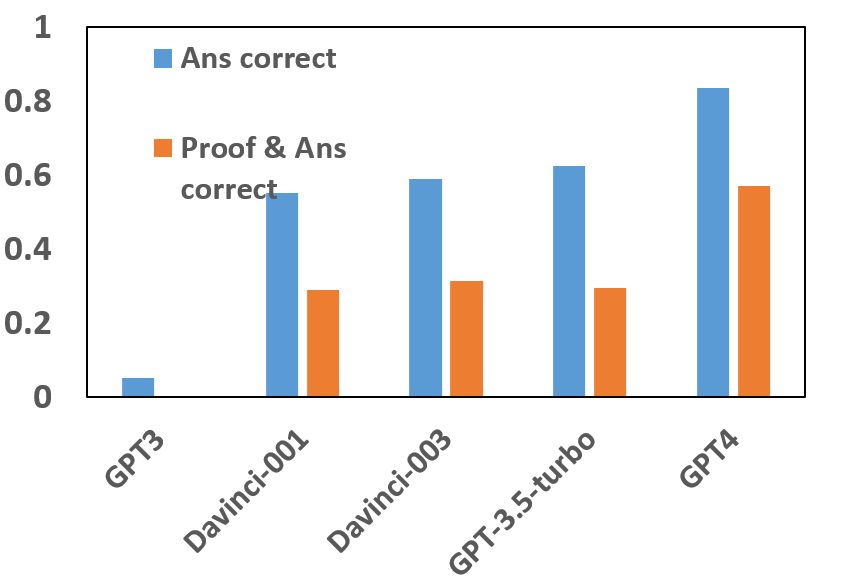}
    \caption{Hallucination rate of LLMs on the relational reasoning task.}
    \label{fig:rr}
\end{figure}

\begin{table}[]
    \centering
    \small
    \begin{tabular}{c|c|c|c}
    \toprule
       Model & $\beta_{\text{num\_theory}}$ & $\beta_{\text{num\_facts}}$ &  $\beta_{\text{num\_arguments}}$   \\
    \midrule
       GPT-3  & \multicolumn{1}{c}{\parbox{1.3cm}{\coloredBar{0}{0}{1}{-}}} & \multicolumn{1}{c}{\parbox{1.3cm}{\coloredBar{0}{0}{1}{-}}} & \multicolumn{1}{c}{\parbox{1.3cm}{\coloredBar{0}{0}{1}{-}}}  \\
       Davinci-001 & \multicolumn{1}{c}{\parbox{1.3cm}{\coloredBar{0.039}{8}{1}{}}} & \multicolumn{1}{c}{\parbox{1.3cm}{\coloredBar{0.068}{8}{1}{$^*$}}} & \multicolumn{1}{c}{\parbox{1.3cm}{\coloredBar{0.052}{8}{1}{}}}  \\
       Davinci-003  & \multicolumn{1}{c}{\parbox{1.3cm}{\coloredBar{0.080}{8}{1}{}}} & \multicolumn{1}{c}{\parbox{1.3cm}{\coloredBar{0.064}{8}{1}{$^*$}}} & \multicolumn{1}{c}{\parbox{1.3cm}{\coloredBar{0.041}{8}{1}{}}} \\
       GPT-3.5-turbo  & \multicolumn{1}{c}{\parbox{1.3cm}{\coloredBar{0.033}{8}{1}{}}} & \multicolumn{1}{c}{\parbox{1.3cm}{\coloredBar{0.037}{8}{1}{$^{**}$}}} & \multicolumn{1}{c}{\parbox{1.3cm}{\coloredBar{0.016}{8}{1}{}}}  \\
       GPT-4  &\multicolumn{1}{c}{\parbox{1.3cm}{\coloredBar{0.20}{8}{1}{$^*$}}}  & \multicolumn{1}{c}{\parbox{1.3cm}{\coloredBar{0.093}{8}{1}{$^*$}}} & \multicolumn{1}{c}{\parbox{1.3cm}{\coloredBar{0.12}{8}{1}{}}} \\
    \bottomrule
    \end{tabular}
    \caption{Regression coefficients for different OpenAI LLMs on relational reasoning task. Statistical significance indicators: ·:p<0.1; *:p<0.05; **:p<0.01; ***: p<0.001.}
    \label{tab:beta-rr}
\end{table}

Figure~\ref{fig:rr} and Table~\ref{tab:beta-rr} shows the hallucination rate of LLMs and the significance of risk factors upon the Ruletaker dataset. From that, we can observe: 

(1) The number of example instances and the model performance is statistically irrelevant (P-value=0.53 for GPT-3.5-turbo). This indicates that the model already understands the task, as increasing the number of examples does not bring in performance improvement. 

(2) Compared to GPT-3, text-davinci-003, GPT-3.5-turbo and GPt-4 demonstrate significantly higher accuracy. These results indicate the pretraining process on the code data can significantly enhance the model's ability to understand complex logical relationships. 

(3) There exists a discrepancy between the result accuracy and the accuracy of both the answer and reasoning process is correct.
This suggests that the LLMs may utilize incorrect intermediate reasoning steps to draw the final conclusions. Hence, it may be inadequate to detect the hallucination level of LLMs only based on results, at least for the relation modeling task. This highlights that when investigating the hallucination of LLMs, especially in the related task, whether or not the generated contents are hallucinated cannot be judged only based on the correctness of results. In the following sections, only the instances with both the answer and reasoning process are correct are taken as not hallucinated.

(4) The hallucination rate of LLMs is significantly related to risk factors about the complexity of relationships, including the number of theories, number of facts, and number of arguments. Take GPT-4 as an example, with the number of theories, number of facts, and number of arguments increasing by 1, the hallucination rate would be $\mathrm{exp}(0.20)=1.22$, $\mathrm{exp}(0.09)=1.09$, and $\mathrm{exp}(0.12)=1.13$ fold, respectively.  

(5) State-of-the-Art LLMs may still struggle with the complex logical reasoning task. For example, GPT-4 has an accuracy of 57\% to generate correct answers based on a correct reasoning process on the Ruletaker dataset. Conclusions (4) and (5) highlight the necessity for incorporating more relational reasoning data into the pretraining and supervised finetuning process of LLMs to further enhance the complex relational modeling ability of LLMs.  

\subsection{Instruction Following}

\begin{table}[]
    \centering
    \small
    \begin{tabular}{c|c}
    \toprule
       Model &  $\beta_{\text{likelihood\_diff}}$ \\
    \midrule
       GPT-3  & \multicolumn{1}{c}{\parbox{4cm}{\coloredBar{0.15}{20}{2}{$*$}}}  \\
       Text-davinci-001 & \multicolumn{1}{c}{\parbox{4cm}{\coloredBar{0.16}{20}{2}{}}} \\
       Text-davinci-003  & \multicolumn{1}{c}{\parbox{4cm}{\coloredBar{0.17}{20}{2}{$**$}}} \\
       GPT-3.5-turbo  & \multicolumn{1}{c}{\parbox{4cm}{\coloredBar{0.015}{20}{2}{$*$}}} \\
       GPT-4  &\multicolumn{1}{c}{\parbox{4cm}{\coloredBar{0.11}{20}{2}{}}} \\
    \bottomrule
    \end{tabular}
    \caption{Hallucination rate of LLMs on the c-NLI task.}
    \label{tab:conflict}
\end{table}

 \begin{figure}
    \centering
    \includegraphics[width=0.3\textwidth]{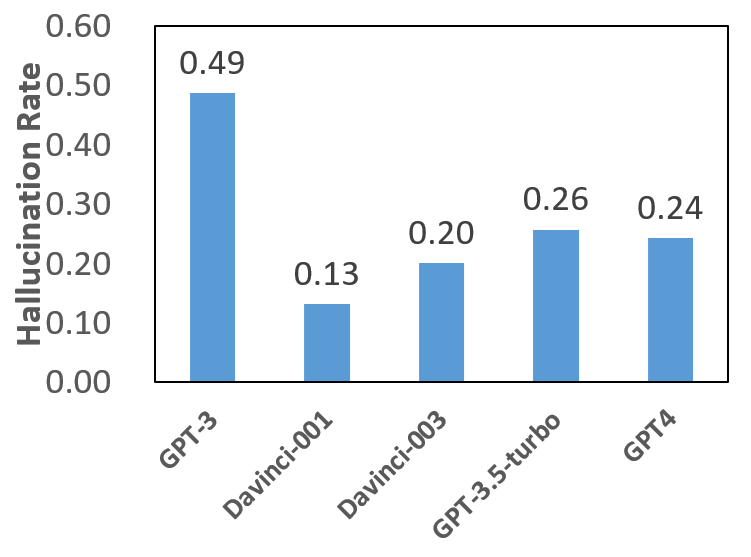}
    \caption{Hallucination rate of LLMs on the c-NLI task.}
    \label{fig:conflict}
\end{figure}


Figure~\ref{fig:conflict} and Table~\ref{tab:conflict} show the hallucination rate of LLMs on the c-NLI task, together with the statistical significance of the decrease of log probability upon the hallucination rate. From which we observe that:

(1) GPT-4 The model performances upon the baseline dataset show that the LLMs can largely understand the corresponding commonsense knowledge.

(2) There is a U-shaped inverse correlation between hallucination rate and model ability: GPT-3, which is not aligned with human cognition, and the state-of-the-art language model GPT-4, both display a relatively high hallucination rate on the c-NLI task. As GPT-3 is not aligned with human beings, it may fail to understand human instructions, so inferences are made only on the probability of language modeling and lead to false results. While for the SoTA LLM GPT-4, note that when the statements in c-NLI are transformed into counterfactual forms, GPT-4 shows a significantly larger likelihood decrease compared to other models. This suggests a stronger language modeling ability, as it can better separate the counterfactual statements by assigning a lower likelihood. Moreover, the decrease in likelihood is significantly positively associated with the hallucination rate. This suggests that the strong language modeling ability may conflict with human instructions, finally leading to hallucinations.
Previously, \citet{mckenzie2023inverse} stated that tasks containing an easy distractor task would lead to inverse scaling. In this paper, we demonstrate one reason for this phenomenon.




\section{Conclusion}

Different from previous works which focus only on certain tasks and quantify the hallucination level of LLMs using solely an index, we aim at providing a new methodology to simultaneously quantify the hallucination level of LLMs and attribute the source of hallucinations by analyzing the association between the probability of hallucination with a set of risk factors. This enables us to unbiasedly measure the hallucination level of LLMs by controlling the effect of potential confounders, and probing the contribution of each risk factor. By recognizing the risk factors through diving into the training process of the LLMs, a set of potential risk factors in the training process are recognized, which offer guidance for further mitigation of the hallucination. Experimental results show that the present SoTA LLMs still suffer from hallucination in key capabilities of LLMs, such as commonsense knowledge memorization, relational reasoning, and instruction following, which highlight the necessity for discovering more risk factors, to facilitate obtaining more trustworthy large language models. Although with a relatively limited number of risk factors, our work can serve as a pioneer to provide insights for future explorations.
\bibliography{anthology}
\end{document}